\documentclass{article}

\usepackage{PRIMEarxiv}

\usepackage[utf8]{inputenc} 
\usepackage[T1]{fontenc}    
\usepackage{hyperref}       
\usepackage{url}            
\usepackage{booktabs}       
\usepackage{amsfonts}       
\usepackage{nicefrac}       
\usepackage{microtype}      
\usepackage{lipsum}
\usepackage{fancyhdr}       
\usepackage{graphicx}       
\graphicspath{{media/}}     
\usepackage{amssymb}
\usepackage{booktabs}
\usepackage{multirow} 
\title{Large Language Models for Automatic Equation Discovery of Nonlinear Dynamics}
\date{}
\author{
  Mengge Du \\
  College of Engineering \\
  Peking University \\
  Beijing \\
   \And
  Yuntian Chen \\
Ningbo Institute of Digital Twin, Eastern Institute of Technology \\
  Ningbo\\
  \texttt{ychen@eitech.edu.cn} \\
  \And
   Zhongzheng Wang \\
  College of Engineering \\
  Peking University \\
  Beijing \\ 
  \And
   Longfeng Nie \\
   School of Environmental Science and Engineering\\
Southern University of Science and Technology\\
  Shenzhen \\ 
  \And
  Dongxiao Zhang \\
  Ningbo Institute of Digital Twin\\
  Eastern Institute of Technology, Ningbo \\
  National Center for Applied Mathematics Shenzhen (NCAMS)\\ Southern University of Science and Technology, Shenzhen \\
  \texttt{dzhang@eitech.edu.cn} \\    
}

\begin{document}
\maketitle
\begin{abstract}
Equation discovery aims to directly extract physical laws from data and has emerged as a pivotal research domain in nonlinear systems. Previous methods based on symbolic mathematics have achieved substantial advancements but often require handcrafted representation rules and complex optimization algorithms. In this paper, we introduce a novel framework that utilizes natural language-based prompts to guide large language models (LLMs) in automatically extracting governing equations from data. Specifically, we first utilize the generation capability of LLMs to generate diverse candidate equations in string form and then evaluate the generated equations based on observations. The best equations are preserved and further refined iteratively using the reasoning capacity of LLMs. We propose two alternately iterated strategies to collaboratively optimize the generated equations. The first strategy uses LLMs as a black-box optimizer to achieve equation self-improvement based on historical samples and their performance. The second strategy instructs LLMs to perform evolutionary operations for a global search. Experiments are conducted on various nonlinear systems described by partial differential equations (PDEs), including Burgers' equation, the Chafee-Infante equation, and the Navier-Stokes equation. Results demonstrate that our framework can discover correct equations that reveal the underlying physical laws. Further comparisons with state-of-the-art models on extensive ordinary differential equations (ODEs) showcase that the equations discovered by our framework possess physical meaning and better generalization capability on unseen data.
\end{abstract}


\section{Introduction}
\label{}
Physical laws often follow concise governing equations, which are crucial for our understanding and transformation of the natural world. With the development of artificial intelligence, simulation of the evolution of nonlinear systems through deep learning has gradually emerged \cite{yeo2019deep,kou2021data,zheng2020purely}. However, these methods are limited by black-box models and lack interpretability. To tackle this issue, equation discovery methods that uncover potential physical laws from observations with explicit mathematical formulas have received increasing attention, which can not only facilitate a deeper understanding of physical processes but also provide domain guidance for data-driven models and enhance their predictive robustness \cite{ISgep,2022pinnreview2}. Moreover, with the governing equation incorporated as physical constraints, neural networks can be equipped with physical intuition and possess better extrapolation ability \cite{lu2021physics,chen_inte}.

In nonlinear systems, states of interest often follow various differential equations, such as ordinary differential equations, in the form of $\dot{\mathbf{x}} = f(\mathbf{x}(t))$, where $\mathbf{x}(t)=\{x_1(t),x_2(t),...,x_n(t)\}^T \in \mathbb{R}^m$ denotes the state variables with the spatial dimension of $m$. The main objective of equation discovery is to find the explicit expression of $f$. Traditionally, this process was based on first principles, which often require experts in the relevant domain to engage in extensive mathematical derivations. In recent years, data-driven methods are gradually rising because of their superior efficiency and applicability \cite{schmidt,2019pnasDD}. In particular, SINDy (Sparse Identification of Nonlinear Dynamics) has emerged as an effective method to tackle this challenge \cite{sindy}. It assumes that the form of $f$ can be simplified as a linear combination of a series of candidate basis functions, where the basis function library is often predetermined based on prior knowledge. With the advantages of high computational efficiency and simple methodology, SINDy has achieved good performance across various fields \cite{sindy_bvp,weak_sindy,sindy_pi,ensemble_sindy}. Nevertheless, the reliance on prior knowledge inherently constrains the applicability of this approach, rendering it challenging to uncover more intricate representational forms. Concurrently, the progress of numerous intelligent optimization algorithms has contributed to the utilization of symbolic mathematics in identifying governing equations with more flexible forms. EQL (Equation Learner) \cite{eql,eql_divide} endeavors to utilize the topological structure of networks to represent equations with different combinations and substitute activations with arithmetic operators, such as $+$ and $-$. An alternative approach seeks to represent equations with expression trees, aiming to discover the optimal equation by optimizing the tree structure. Common optimization methods are based on gradient descent \cite{kamienny2023generativesr,valipour2021symbolicgpt,li2022transformer}, reinforcement learning \cite{dsr,sun2022symbolic,PRRDISCOVER,sga_pde,Rdiscover}, or evolutionary algorithms \cite{ISgenetic,gp_sr2,gp_sr3}. These approaches substantially diminish the reliance on prior physical knowledge, enabling wider application scenarios. However, laborious and intricate algorithm design and coding efforts are required for equation generation and optimization, which is not conducive to wide-scale promotion.

Transformer-based large language models (LLMs) have continuously emerged and have achieved remarkable results in various application domains in recent years \cite{LLM_utilize1,LLM_utilize2,LLM_utilize3}. A vast number of trainable parameters and a large diverse training corpus enable LLMs to possess strong generation and reasoning capabilities. Some recent studies have started to explore the potential of LLMs in mathematical reasoning \cite{LLM_mathematicaldiscover}, algorithmic optimization \cite{AEL_evo}, and code generation \cite{wang2023codet5+}, with some even employing LLMs as direct optimizers to tackle black-box optimization challenges \cite{liullm_es_optimizer}. A salient question is whether we can leverage LLMs to automatically complete equation discovery without additional parametric models and optimization processes.

In this paper, we propose a LLM-based framework for automatic equation discovery, as shown in Fig.~\ref{fig:overview}. Initial equations are first generated in string format after prompting LLMs with a clear symbol library and problem descriptions. The equations can be seamlessly parsed and transformed into expression trees via the domain tool in symbolic mathematics and evaluated based on the score function and data. Elite equations are preserved in the priority queue and incorporated into prompts to guide iterative optimization by LLMs. During the optimization phase, LLMs can serve as an optimizer to conduct the self-improvement process. Some local refinements are applied to the historical equations based on the analysis of the inherent relationship between the combinations of symbols and their performance. In addition, well-designed prompts are used to guide LLMs to apply user-defined evolution operators on elite equations, promoting the generation of more diverse equation combinations. These two approaches are iteratively employed in an alternating manner to refine the structure of generated equations until the optimal equation satisfies the termination conditions.
Our framework has been tested for uncovering the correct PDE equations in several canonical nonlinear systems and has verified that the two optimization approaches of local modification and random evolution have a synergistic effect. In addition, we further validated our framework on sixteen one-dimensional ODE systems, and the results showed that it could achieve comparable performance to the state-of-the-art and have better generalization capabilities. Our main contributions are as follows:
\begin{itemize}
\item We propose an automated equation discovery framework that utilizes the natural language generation and reasoning capabilities of LLMs. The framework eliminates the need for manually crafting intricate programs for equation generators and optimizers and is totally parametric-free during optimization.
\item We employ manually designed prompts to guide LLMs in executing two optimization approaches: self-improvement and evolutionary search. The alternating iterative optimization strategy effectively strikes a balance between exploration and exploitation.
\item We validate the efficacy of our framework through a series of experiments on ODEs and PDEs. The results demonstrate that its performance is on par with or even better than the state-of-the-art symbolic regression (SR) methods, especially in generalization capability. The framework encourages more extensive research and application of LLMs in the domain of equation discovery.
\end{itemize}
\section{Related Works}
\subsection{Symbolic Equation discovery}
Symbolic mathematics-based methods can directly uncover the potential relationships between variables from data. With the development of computational equipment and machine learning, these methods have gradually gained increasing attention. Equation discovery tasks typically encompass three phases: generation, evaluation, and optimization. In the generation stage, based on certain context-free s~\cite{ISbrence2023,cfg_parkes2008concise}, equations in mathematical form are typically transformed into expression trees. The internal nodes of the expression tree are predefined operators (e.g., $+, -$) and operands (e.g., observations $x$ or constant). By conducting a top-down traversal of the expression, a unique sequential representation can be generated. This representation is more concise and enables more efficient batch generation and gradient-based optimization~\cite{dsr,dsr_gp}. Some constraints are carefully designed to generate dimensional consistent expressions and ensure the physical and mathematical rationality. In the evaluation stage, the main focus is to assess the performance of the discovered equations in terms of their fit to the data and complexity. Finally, in the optimization stage, the commonly utilized algorithms mainly include genetic programming~\cite{gp_sr4}, gradient descent-based neural network models~\cite{eql}, and recently emerging reinforcement learning models \cite{dsr,sun2022symbolic}. At the same time, pretrained models based on transformers have gradually emerged \cite{biggio2021neural,valipour2021symbolicgpt,vastl2024symformer,odeformer}. These models are trained on a large amount of data and can directly output the discovered equation results based on the observations, greatly accelerating the inference speed. Evidently, approaches founded on symbolic mathematics necessitate manually designed algorithms in multiple aspects, elevating the learning and application barrier. Conversely, our framework, guided by natural language, significantly streamlines the generation and optimization components, enabling researchers to concentrate solely on the evaluation aspect, where domain expertise is genuinely essential.

\subsection{Large language model for optimization}
The powerful language understanding and generation capabilities of large models have led to their extensive application in various fields~\cite{LLM_utilize1,LLM_utilize2,LLM_utilize3,LLM_utilize4}. Studies have recently demonstrated the feasibility of employing prompt engineering to direct LLMs in addressing optimization problems. One approach is to directly use LLMs as optimizers in a self-improvement updating manner \cite{Yang_selfoptimize,guo_selfoptimize}. Taking into account the problem definition and previously generated solutions, LLMs can be directed to refine candidate solutions iteratively.  The findings suggest that LLMs possess the capability to progressively improve the generated solutions by building upon the knowledge gained from past optimization results. Other related works attempt to combine LLMs with evolutionary search methods to solve optimization problems. Prompts can be designed to instruct LLMs to execute evolutionary algorithms to incrementally enhance the existing solutions within the population. This synergistic combination ultimately leads to the discovery of novel insights and advancements in addressing open research questions, including combinatorial optimization problems like (e.g. traveling salesman problems \cite{liullm_es_optimizer}), multiobjective evolutionary optimization \cite{liufeiMOEO_LLM}, prompt optimization \cite{promot_evo}, algorithm design \cite{AEL_evo,LLM_mathematicaldiscover}, game design \cite{Lan_LLMea_game}, and evolutionary strategies \cite{LLM4ES}.

Our method pioneered the application of LLMs in the field of equation discovery, constructing a plug-and-play discovery framework. By leveraging natural language, we have seamlessly integrated the self-improvement capabilities of LLMs with evolutionary search techniques, which effectively strikes a balance between exploitation and exploration. The proposed method ensures the stability and efficiency of optimization while finding the globally optimal equation.

\begin{figure}[!ht]
\includegraphics[width=\textwidth]{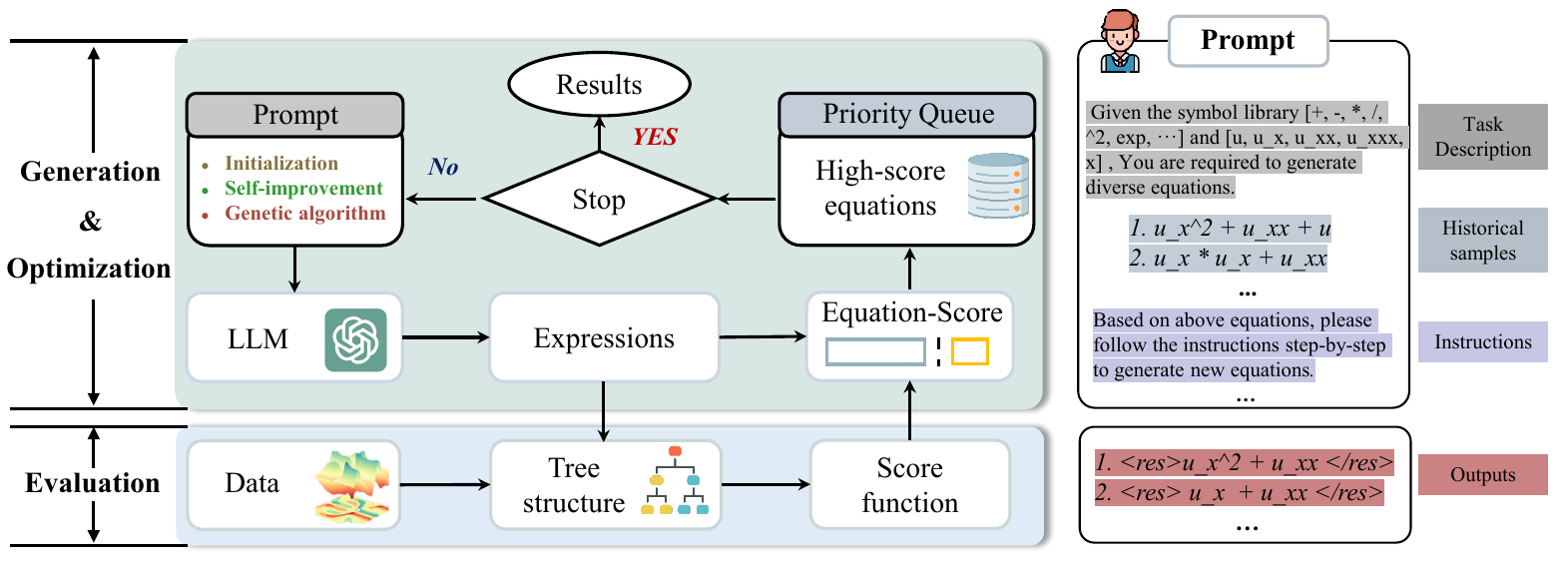}
\caption{ 
Overview of the proposed framework.}
\label{fig:overview}
\end{figure}

\section{Methods}
\subsection{Problem overview}
The goal of the equation discovery task is to identify an explicit mathematical expression $\mathcal{F}$, defined by mathematical symbols, based on a given set of observations. The true form $\mathcal{F}$ should satisfy 
$$
\dot{x}=\mathcal{F}(x;\xi), \quad \mathcal{F}: \mathbb{R}^D \rightarrow \mathbb{R}
$$
where the state variable $x(t) \in \mathbb{R}^D$; $\dot{x}$ refers to the time derivatives; and $\xi$ denotes the possible constants. We aim to find an optimal expression $\mathcal{F}$ that accurately describes the true underlying physical laws in the dynamical system while keeping the form concise.
The form of $\mathcal{F}$ may differ slightly for nonlinear systems governed by different types of equations. In this paper, we consider two types of governing equations: PDEs and ODEs. For ODEs, the form of equations can be generated by freely combining symbols from a predefined library, including constants $\xi$. The value of $\xi$ is typically determined using optimization techniques that minimize a specific data fitting metric, such as the mean square error (MSE).
For PDEs, the right-hand side of the equation often consists of the combinations of state variables (e.g., $u$) and their spatial derivatives (e.g., $u_x$ and $u_{xx}$). For example, the Burgers' equation is represented as $u_t=auu_x+bu_{xx}$. Similar to the previous SINDy-based methods \cite{sindy,pde_find}, we simplify $\mathcal{F}$ to be represented by a linear combination of a series of basis function terms $\Theta(u,x)$. The difference is that function terms can be represented by any combination of symbols without constants rather than the predefined monomials. Constants only appear as coefficients of the function terms, i.e., $\mathcal{F}\approx\Theta(u, x) \cdot \xi$. The coefficients $\xi$ of the function terms can then be obtained through sparse regression.

In this paper, the skeleton of $\mathcal{F}$ is generated and refined by LLMs. We can further empower LLMs with SymPy \cite{meurer2017sympy}, a domain-specific, open-source Python library for symbolic mathematics, to parse string-form equations and convert them into expression trees, which facilitates the evaluation of data fitting.

\subsection{Framework}
\begin{figure}[!ht]
\centering
\includegraphics[width=0.6\textwidth]{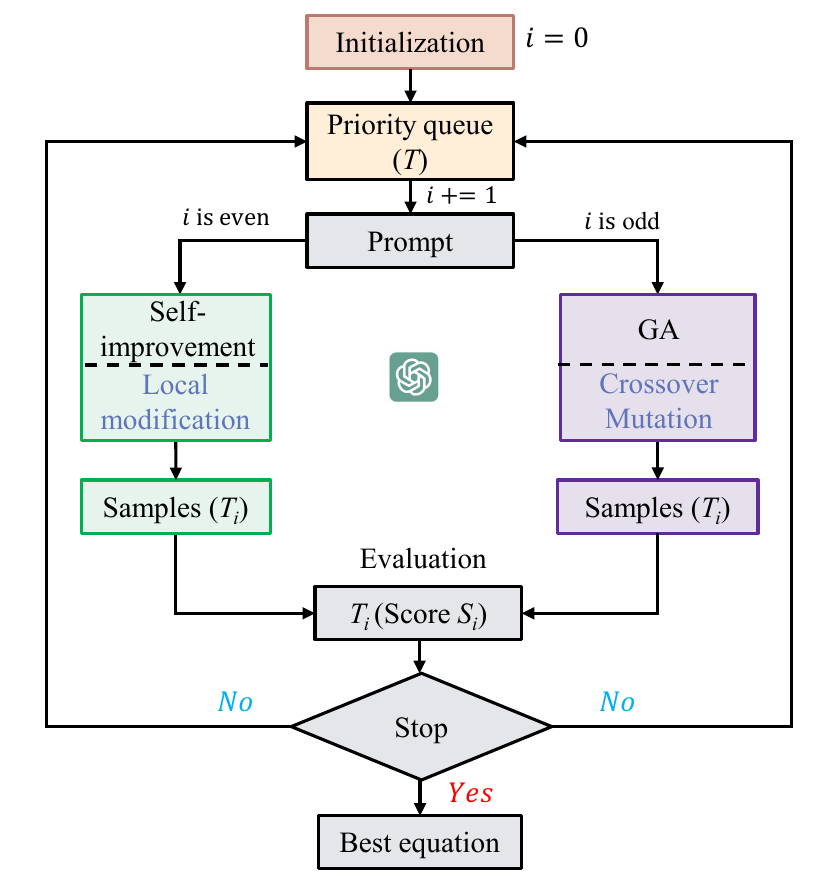}
\caption{
Workflow of the proposed framework. }
\label{fig:procedures}
\end{figure}

Our framework employs natural language to guide LLMs in generating and refining equations, which is shown in Fig.~\ref{fig:procedures}. First, LLMs draw upon extensive prior training data, and they tend to produce mathematically reasonable expressions with concrete descriptions of the physical process. Second, we employ an alternating iterative approach that combines self-improvement and evolutionary search to refine the generated equations. Users are only required to concentrate on establishing appropriate evaluation criteria, i.e., the score function, to precisely evaluate the generated equations. Equations with higher scores are used to update a priority queue that retains the top $K$ optimal samples up to the current iteration, while bad samples are discarded. These elite equations are preserved in equation-score format and can be incorporated as in-context examples in prompts to instruct LLMs to generate better-fitting equations. The specific components and procedures of the framework are described below in detail.

\subsection{Prompt Engineering}

Throughout the entire equation discovery process, the generation and optimization process are instructed with natural language-based prompts, which follow a unified structure in the three processes of initialization, evolution, and self-improvement. The standard format consists of the following components, as shown in Fig.~\ref{fig:overview}.
\begin{itemize}
\item Task descriptions: This part primarily explains the scientific task and defines the symbol library, including operators (e.g., {$+,-$} and operands (e.g., {$x,const$}). If the mechanism of the entire physical process is understood, the physical meanings of different variables can also be provided in this context. This is essential for producing effective and reasonable equations and can significantly reduce the search space.
\item Historical examples: To guide LLMs to generate better-fitting equations, $M$ high-quality equations from past iterations are incorporated as in-context examples in the prompts. The selected equations are chosen from two sources. First, all of the $K$ expressions within the priority queue are used for stable and efficient optimization. Second, we select $M-K$ expressions from the last iteration to maintain sampling diversity. Notably, the presentation of these samples within the prompt varies according to the optimization techniques employed. In the evolution process, only high-quality equations in string format are presented, while in the self-improvement process, historical samples are shown in the form of equation-score pairs.
\item Instructions: This part is primarily aimed at guiding LLMs to generate and refine equations. In the initialization stage, LLMs are required to freely combine symbols from the library to produce equations of arbitrary form and length. In the optimization stage, LLMs are mainly guided to generate refined equations based on historical equations according to different optimization strategies. For the self-improvement process, how to conduct the local modifications should be clearly introduced. The instructions in the LLM-guided evolutionary search should emphasize the concrete implementation details of the selection and evolution of equations.
\item Other hints or constraints: If the relationship of physical variables is available, we can directly describe the requirements for the structure of the generated equations through natural language. In the optimization stage, we can further define local modifications and evolutionary operators and provide possible examples as few-shot prompts. In addition, some hints about the format of outputs can be incorporated here.

\end{itemize}
The utilized prompts in this paper are demonstrated in Appendix \ref{append:prompt}.

\subsection{Initialization}
The initial equation population can be generated through LLMs or based on prior knowledge, i.e., manually predefined equations. In this study, we employ prompts to direct LLMs in randomly generating the initial population with a given symbol library and problem descriptions. First, LLMs have been trained on extensive text data, enabling them to learn numerous effective equation representations. Consequently, the generated equations generally follow mathematical principles. Second, constraints can be established using natural language, thereby preventing the occurrence of equations that violate the specified conditions. For instance, constraints can include restricting the equation length, frequency of specific symbols, and preventing the generation of invalid nested combinations. Traditionally, implementing these constraints necessitated intricate code, such as generating equations based on probabilistic context-free s \cite{ISbrence2023,BRENCE_PROGED} or subtly modifying probabilities during the symbol sampling process \cite{dsr}.

\subsection{Evaluation}
LLMs excels at creative generation based on enormous corpus, but need to be further strengthened with domain tools and human-designed feedback to deal with the symbolic discovery task.   Regarding the equation skeletons that have been generated in string format, we can employ Sympy \cite{meurer2017sympy} to parse and instantiate them as corresponding symbolic expression trees. Before evaluating them, we first need to determine the parameters in the expressions, i.e., constants, and then further score them according to the designed score function.

\subsubsection{Constant optimization}
\begin{figure}[!ht]
\centering
\includegraphics[width=\textwidth]{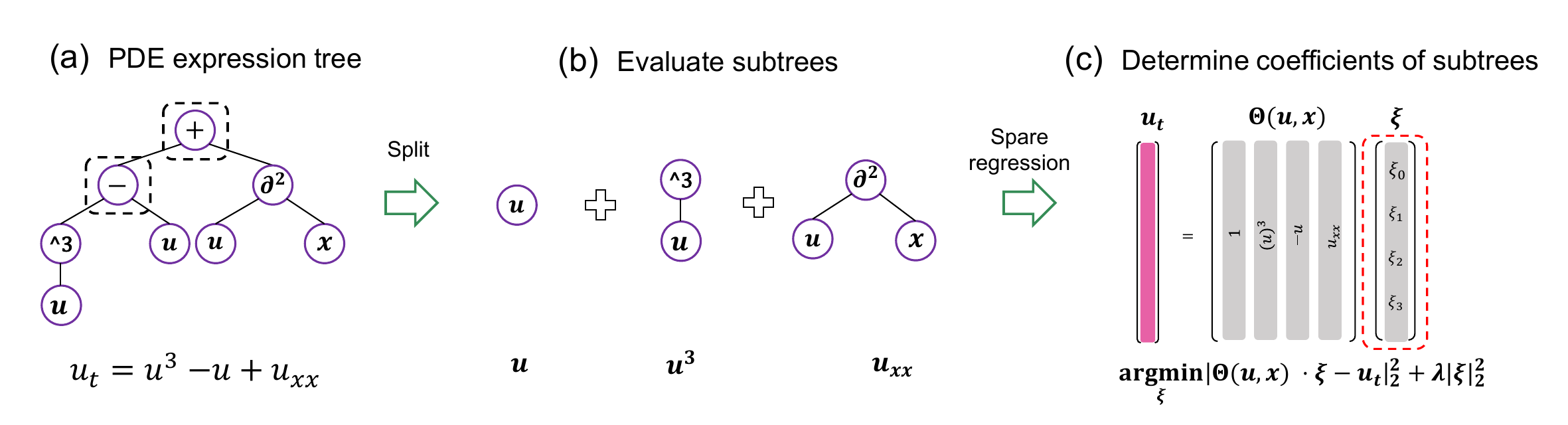}
\caption{
Determination of PDE coefficients. }
\label{fig:coefficients}
\end{figure}

This study considers two types of governing equations: PDEs and ODEs. Depending on the specific features of the equations they represent, we adopt two distinct approaches to evaluate the constants. For PDEs, constants mainly appear as coefficients of function terms. Therefore, we first need to decompose the expression tree by splitting it into equation terms based on the "+" and "-" operators at the top of the tree, and then further solve for the coefficients using sparse regression methods, as shown in Fig.~\ref{fig:coefficients}. Terms with nontrivial coefficients will be kept and others will be removed for simplicity. 
\begin{equation}
\xi^*_{pde}=\arg \min _{\xi}\left|\Theta(u,x) \cdot \xi-u_t\right|_2^2+\lambda|\xi|_2^2
\end{equation}
For ODEs, constants can appear at any position in the expression tree. We first generate equation skeleton through LLMs, and then utilize the Broyden-Fletcher-Goldfarb-Shanno algorithm (BFGS) \cite{bfgs} to execute the following optimization objective.
\begin{equation}
\xi^*_{ode} = argmin_{\xi} \sum_{i=1}^n \frac{1}{n} (\dot{x}-\mathcal{F}(x_i;\xi))^2 
\end{equation}
Rounds of optimization iterations are performed using scipy.optimize.minimize to ultimately determine all the constants in the expression tree. Note that if the generated equations do not contain constant operands, sparse regression techniques can be employed to assign coefficients other than 1 to each term, thereby improving the accuracy of the discovered equations and facilitating the identification of lengthy true equations.

\subsubsection{Score function}
After obtaining the values of the constants in the equation, we designed a score function to evaluate the performance of the generated equations.
\begin{equation}
S=\frac{1-\zeta_1 \times m }{1+NRMSE}
\end{equation}
\begin{equation}
NRMSE=\frac{1}{\sigma_{\dot{x}}}\sqrt{\frac{1}{N} \sum_{i=1}^N\left(\dot{x}_{t_i}-\mathcal{F}(x_i)\right)^2}
\end{equation}
where the normalized root-mean-square error (NRMSE) is employed as a fitness metric to evaluate the discrepancy between the left and right sides of the equation. We penalize the number of equation terms $m$ in the equation numerator to encourage finding more concise forms and $\zeta$ refers to the penalty coefficient.
Through the designed score function, we can assign a score to each equation, then select elite equations, and introduce them into the prompt to guide subsequent optimization.

\subsection{Optimization}
This study utilizes two LLM-guided optimization techniques to enhance the equation refinement process.  The self-improvement method primarily performs local modifications based on the equation's performance, while the genetic algorithm-based approach is employed for a global search on the elite equations. Our goal is to achieve a better balance between exploration and exploitation.

\subsubsection{Self-improvement process}
LLMs have been demonstrated in numerous experiments to function as gradient-free optimizers, possessing the ability to draw inferences from historical data and iteratively optimize to produce superior samples~\cite{liullm_es_optimizer}. We include historical elite equations and their corresponding scores as equation-score pairs within the prompt, enabling LLMs to perform local modifications using these data. The modifications primarily encompass two facets: (1) recognizing and eliminating redundant equation terms by leveraging historical data; (2) incorporating and generating novel random equation terms built upon existing equations. These two operations resemble the introduction of "delete" and "add" operators, which can effectively utilize the historical elite samples and aptly supplement the potentially unstable updating of genetic algorithms. An example case of the self-improvement process is shown in Fig.~\ref{fig:selfimpro} and the customized prompts are demonstrated in Appendix~\ref{append:prompt}.

\begin{figure}[!ht]
\centering
\includegraphics[width=0.6\textwidth]{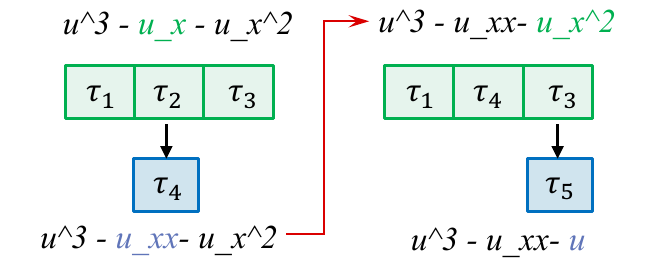}
\caption{
Self-improvement process executed by LLMs.}
\label{fig:selfimpro}
\end{figure}

\subsubsection{Equation evolution process}
Genetic algorithms are one of the commonly used global optimization methods inspired by natural selection \cite{ga1,ga2}. Evolutionary operators can be applied to the parent individuals to generate new offspring. In particular, this procedure requires an intricate design and application on tree structures in symbolic regression. In this paper, we employ natural language to guide LLMs in the execution of the genetic algorithms, rather than relying on manual coding. Specifically, we conduct crossover and mutation operations on the $M$ equation populations generated in the past, thereby producing a greater variety of equation combinations. The process consists of two steps:

\paragraph{Construct parent population} Historical elite equations will be incorporated into the prompt for the evolution process, originating from two sources:  a predefined priority queue caching the top $K$ historically elite equations and high-quality samples selected from the last iteration. By combining them, we ultimately retain the $M$ better-performing equations as the parent population.

\begin{figure}[!ht]
\centering
\includegraphics[width=0.7\textwidth]{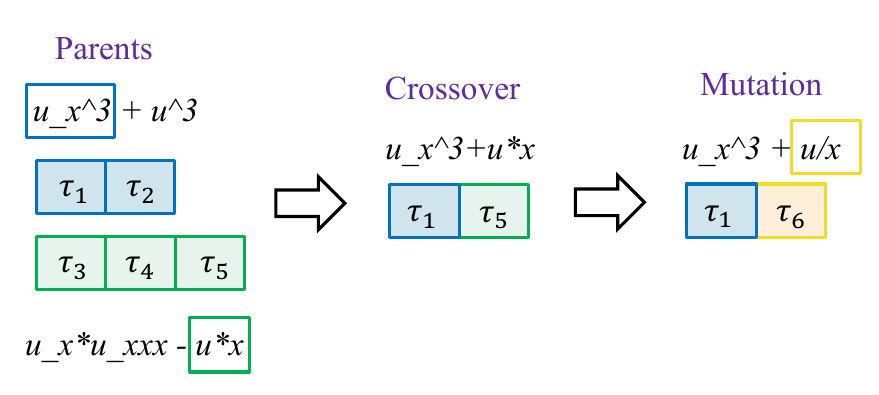}
\caption{
Crossover and mutation executed by LLMs. }
\label{fig:evol}
\end{figure}

\paragraph{Selection and evolution} The entire process comprises three steps. First, LLMs randomly select two equations from the population as parents and then guide them to perform equation crossover to produce new equations. This process can involve both the crossover of entire equation terms and the crossover within equation terms. Finally, further mutations of operands or operators are performed based on the new equations.  Ultimately, iterating the three steps until $M$ offspring are produced. The whole process is directed and performed in natural language and is shown schematically in Fig.~\ref{fig:evol}.

Self-improvement based on local modifications effectively utilizes the reasoning capabilities of LLMs to identify the direct mapping relationships between symbol combinations and scores, thereby refining equations in a gradient-free manner. However, this approach is prone to becoming trapped in local optima. On the other hand, genetic algorithms exhibit strong global optimization abilities, but their effectiveness heavily depends on the quality of the initial population, resulting in possible updating instability. By integrating these two strategies and employing a dynamic buffer of high-quality equations, the evolution and refinement process of equations can be significantly complemented and enhanced.

\section{Resutls}
\subsection{Evaluation metrics}
The experimental section provides the discovery results of the suggested framework for both PDEs and ODEs. We consider PDE equations to be represented as a linear combination of equation terms of arbitrary form, and the constants are primarily solved through sparse regression. Our goal is to find the exact equation form, and the accuracy of the identified equation is assessed by determining the equation's coefficient error.
$$
E=\frac{1}{n} \sum_{i=1}^n \frac{\left|\xi_i^*-\xi_i\right|}{\left|\xi_i\right|} \times 100 \%
$$
where $n$ denotes the total number of function terms; $\xi_i$, $\xi_i^*$ refer to the true coefficients and identified coefficients, respectively.
ODEs are more complex in symbolic form. A skeleton with defined symbols needs to be constructed first, followed by the optimization of constants within the skeleton, which may generate more symbol combinations. Compared to identifying the most consistent expression in symbolic form, it is more crucial and meaningful to conduct numerical evaluation. Specifically, we aim to find an effective $\hat \mathcal{F}$, whose solution trajectories approximate the observed $x$ in the current numerical domain, i.e., all of the expressions are evaluated by the reconstruction accuracy. Furthermore, the other critical criterion is that the solution of the identified $\hat \mathcal{F}$ precisely fits the correct trajectories even when the initial condition varies. We utilize the coefficient of determination ($R^2$) as the metric to evaluate the agreement between the solution trajectories and the true trajectories: $R^2=1-\frac{\sum_{i}^{n}\left(x_i-\hat{x}_i\right)^2}{\sum_{i}^{n}\left(x_i-\bar{x}\right)^2} \in(-\infty, 1]$, where $x_i$ denotes observations and $\hat{x}_i$ refers to predicted values.

\subsection{Experiment settings}
The hyperparameters used in the experiments are shown in Table~\ref{table:hyperp}. GPT-3.5-turbo is utilized as the default LLM backbone. In terms of the experimental setup, the symbol library and equation assumptions used for mining ODEs and PDEs are slightly different, as shown in Table~\ref{table:setting}. The library used for mining PDEs has relatively fewer operators and more operands involved and does not include the symbol "const". On the other hand, the library used for mining ODEs covers more mathematical operators, and constants are determined using nonlinear optimization methods, e.g., BFGS~\cite{bfgs}.

\begin{table*}
  \caption{Default hyperparameter settings.}
  \label{table:hyperp}
  \centering
\begin{tabular}{ccc}
\hline
Hyperparameter &  Default value &  Definition                           \\ \hline
$M$     & 10             & Number of expressions generated at each iteration       \\
$P$     & 100             & Number of total iterations       \\ 
$K$     & 5            & Size of the priority queue     \\ 
$N_{term}$     & 6             & Maximum number of function terms               \\ 
$\zeta_1$           & 0.01          & Parsimony penalty factor for redundant function terms             \\
$\lambda$           & 0.001             & Weight of the STRidge regularization term    \\
$T$           & 0.9       & LLM temperature

\\ \hline
\end{tabular}
\end{table*}

\begin{table}[hb]
\caption{Default experimental settings for discovering different systems.}
  \label{table:setting}
  \centering
 \resizebox{\textwidth}{!}{
\begin{tabular}{cccc}
\hline
Nonlinear Systems & Operators & Operands & Constants optimization \\ \hline
ODE               & {$+,-,\times,\div, \land ,sin,cos,log,exp$}        & {$x,const$}      & Nonlinear \\
PDE               & {$+,-,\times,\div, \land^2, \land^3$}        & {$u,x,u_x,u_{xx},u_{xxx}, u_{xxxx}$}       & Linear \\ \hline
\end{tabular}
}
\end{table}

\subsection{PDE discovery task}
\subsubsection{Equations and discovered results}

\begin{table}[!ht]
\caption{Summary of canonical nonlinear systems governed by PDEs and discovered results. The subscripts $m$ and $n$ denote the number of discretizations.}
  \label{table:PDE_table}
  \centering
 \renewcommand{\arraystretch}{2}
 \resizebox{\textwidth}{!}{
\begin{tabular}{lccc}
\hline
\multicolumn{1}{c}{PDE systems} & Form & Coefficient error & Data discretization        \\ \midrule
    \begin{minipage}{.2\textwidth}
      \includegraphics[width=\linewidth, height=20mm]{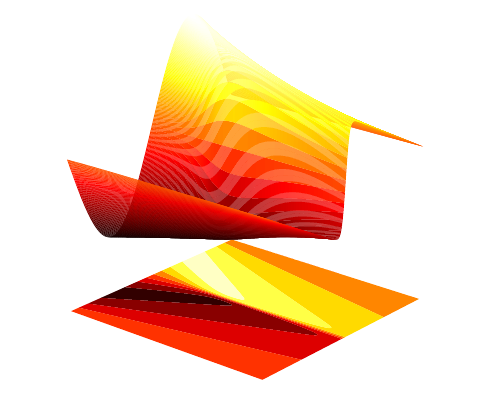}
    \end{minipage}  Burgers       
&$u_t=-uu_x+0.1u_{xx}$         &$1.25\pm1.63\%$ & $x \in[-8,8)_{m
=256}, t \in[0,10]_{n=201}$               \\
    \begin{minipage}{.2\textwidth}
      \includegraphics[width=\linewidth, height=20mm]{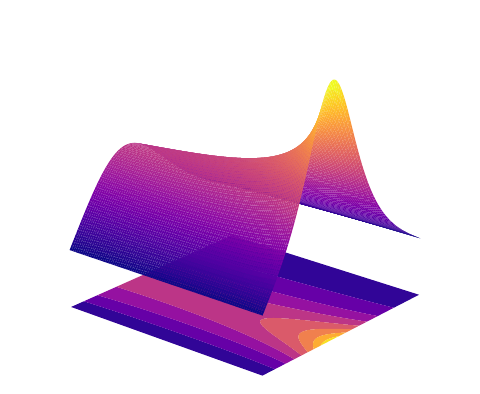}
    \end{minipage} Chafee-Infante
&$u_t=u_{xx}+u-u^3$          &$0.05\pm0.03\%$ &  $x \in[0,3]_{m
=301}, t \in[0,0.5]_{n=200}$       \\
    \begin{minipage}{.2\textwidth}
      \includegraphics[width=\linewidth, height=20mm]{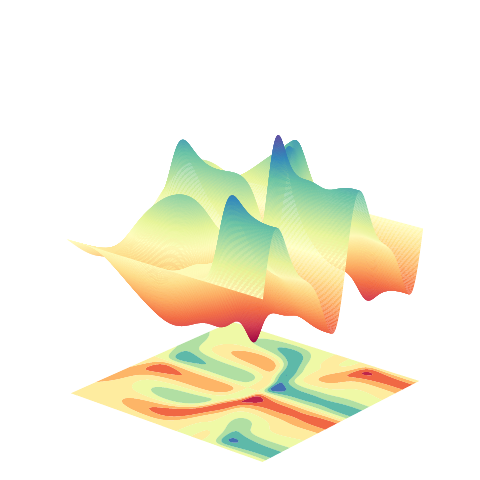}
    \end{minipage}  KS
& $u_t=-uu_x-u_{xx}-u_{xxxx}$  & $0.5\pm0.2\%$        & $x \in[-10,10]_{m
=512}, t \in[0,20]_{n=256}$    \\
    \begin{minipage}{.2\textwidth}
      \includegraphics[width=\linewidth, height=20mm]{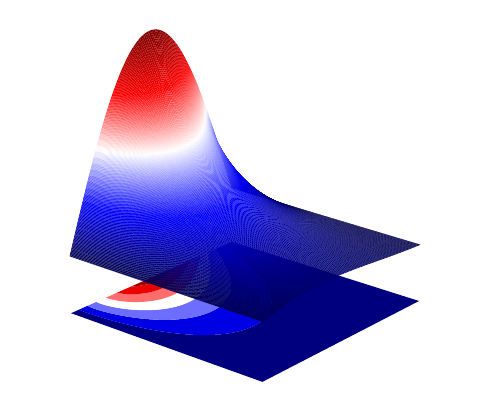}
    \end{minipage} PDE\textunderscore divide
&  $u_t=-u_x/x+0.25u_{xx}$            & $0.15\pm0.09\%$  &$x \in[1,2)_{m
=100}, t \in[0,1]_{n=251}$\\
    \begin{minipage}{.2\textwidth}
      \includegraphics[width=\linewidth, height=20mm]{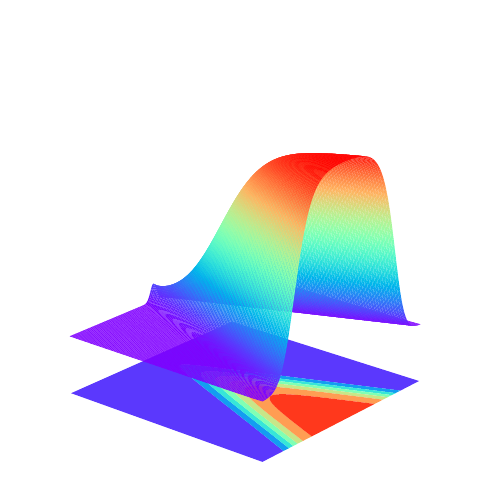}
    \end{minipage} Fisher-KPP     
&  $u_t=0.02uu_{xx}+0.02(u_x)^2+10u-10u^2$            & $1.34\pm0.38\%$  &$x \in(-1,1)_{m
=199}, t \in(0,1)_{n=99}$ \\
 
    \begin{minipage}{.2\textwidth}
      \includegraphics[width=\linewidth, height=18mm]{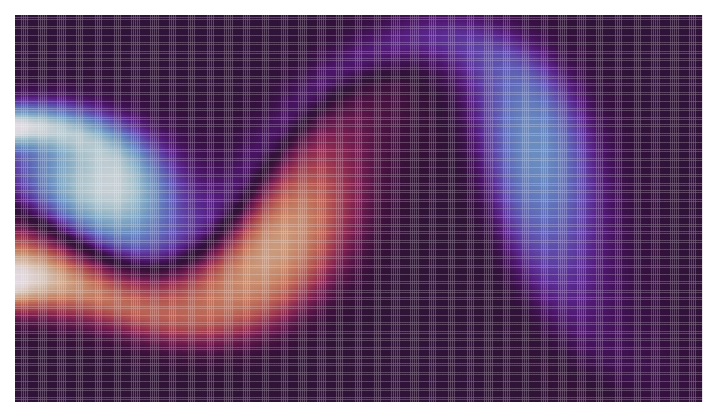}
    \end{minipage} NS
&  $\omega_t=0.1\omega_{xx}+0.1\omega_{yy}-u\omega_x-v\omega_y$            & $0.15\pm0.09\%$  &\begin{tabular}[c]{@{}c@{}}$x \in[0,6.5]_{m=325}, y \in[0,3.4]_{my=325}$, \\ $t \in[0,30]_{n=150}$\end{tabular}  \\ 

\hline
\end{tabular}
}
\end{table}

\begin{figure}[!ht]
\centering
\includegraphics[width=\textwidth]{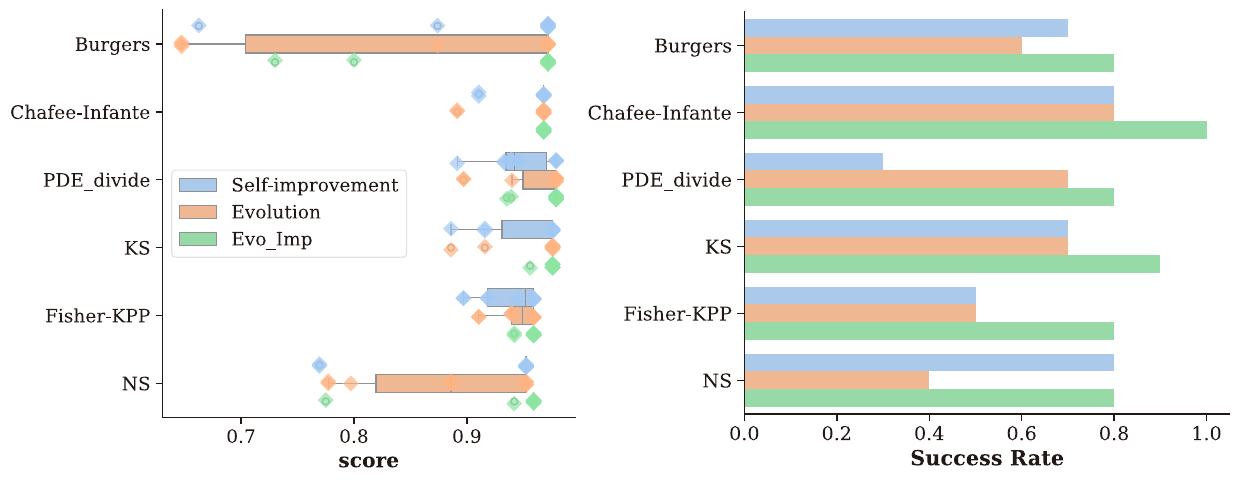}
\caption{
Discovered results under different optimization methods.}
\label{fig:pde_sum}
\end{figure}
In the experiments of PDE discovery, we demonstrate the framework’s ability to discover the governing equations of six canonical nonlinear systems, including the Burges’ equation, Chafee-Infante equation, PDE\_divide equation with fractional structure, Kuramoto-Sivashinsky (KS) equation with fourth-order derivatives, nonlinear Fisher–Kolmogorov–Petrovsky–Piskunov (Fisher-KPP) equations with a square of the spatial derivative, and two-dimensional Navier-Stokes (NS) equation. With the default parameter configuration, our approach accurately identifies the correct structure of the equations while maintaining minimal coefficient error, as shown in Fig.~\ref{table:PDE_table}. Notably, in comparison to fixed candidate set methods, our framework reduces the dependence on prior knowledge, enabling the discovery of more complex equation forms, such as equations with fractional or compound structures.

\subsubsection{Comparison of different optimization strategies}
We further verified the effectiveness of the proposed LLM-guided iterative optimization. Three optimization methods are primarily discussed and compared: (1) using only the self-improvement optimization method; (2) using only the genetic algorithm; and (3) the alternating iterative method combining the two methods (as proposed in the framework). The identification experiments for the aforementioned equations were replicated ten times, with each experiment's maximum iteration count set to 50, to further examine the efficiency of various methods. Fig.~\ref{fig:pde_sum} illustrates that the iterative approach combining both methods yielded the highest frequency of discovering the correct equations, with recovery rates consistently surpassing 80\%, outperforming the outcomes achieved by employing a single optimization technique. Fig.~\ref{fig:pde_sum}(b) depicts the success rate of the ultimately identified equations. It is noteworthy that despite the self-improvement method outperforming the genetic algorithm and exhibiting higher optimization efficiency in some systems, such as the Burgers' equation, it is more prone to converging to local optima. When the iteration count is extended to 100 steps, the symbolic success rate of optimization employing the genetic algorithm approach surpasses 80\% for all of the equations, demonstrating its superior global optimization capability, whereas self-improvement hardly achieves significant improvement.

\begin{figure}[!ht]
\centering
\includegraphics[width=\textwidth]{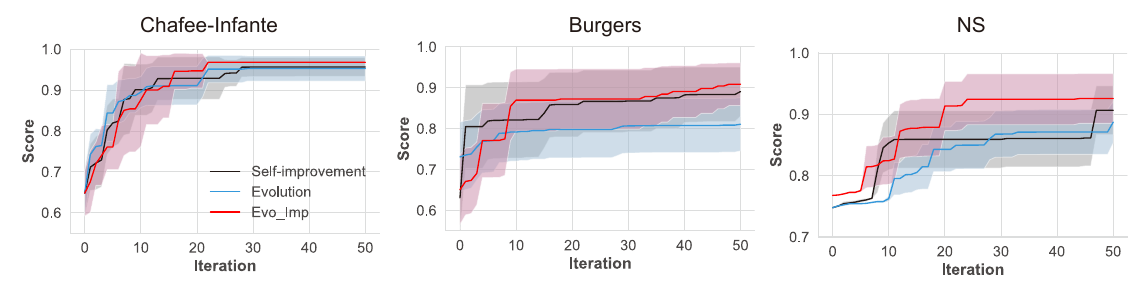}
\caption{
Evolution of the maximum score with different methods while discovering the Chafee-Infante equation, Burgers' equation and NS equation.}
\label{fig:reward_compare}
\end{figure}

\begin{figure}[!ht]
\centering
\includegraphics[width=\textwidth]{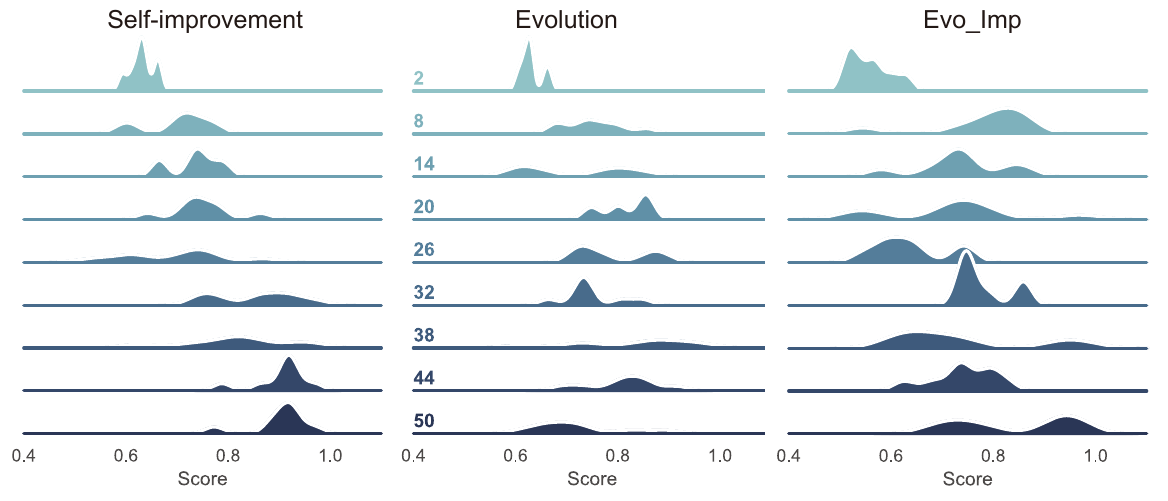}
\caption{
Distribution of scores at different iterations with different methods while discovering the Chafee-Infante equation. Numbers in the figure refer to different iteration steps.}
\label{fig:ridge_plot}
\end{figure}

We provide further detailed analysis with the Chafee-Infante, Burgers, and NS equations utilized as examples. Fig.~\ref{fig:reward_compare} illustrates the evolution of the maximum score throughout the optimization process. It reveals that the optimization efficiency of the alternating approach combining both methods is superior, facilitating faster identification of the correct equations in various equation discovery tasks. Utilizing the Chafee-Infante equation as an illustration, we further examine the density distribution of scores at each iteration throughout the optimization process. Fig.~\ref{fig:ridge_plot} illustrates that the self-improvement strategy exhibits a propensity for local modifications on historical elite equations, resulting in the overall score resembling an incremental trend. Conversely, GA excel at global searches, identifying equations with higher diversity, albeit at the cost of potentially compromised optimization efficiency. Employing alternating iterations of both methods proves more advantageous in striking a balance between exploration and exploitation.

\subsection{ODE discovery task}

\subsubsection{Discovered results}
\begin{figure}[!ht]
\centering
\includegraphics[width=\textwidth]{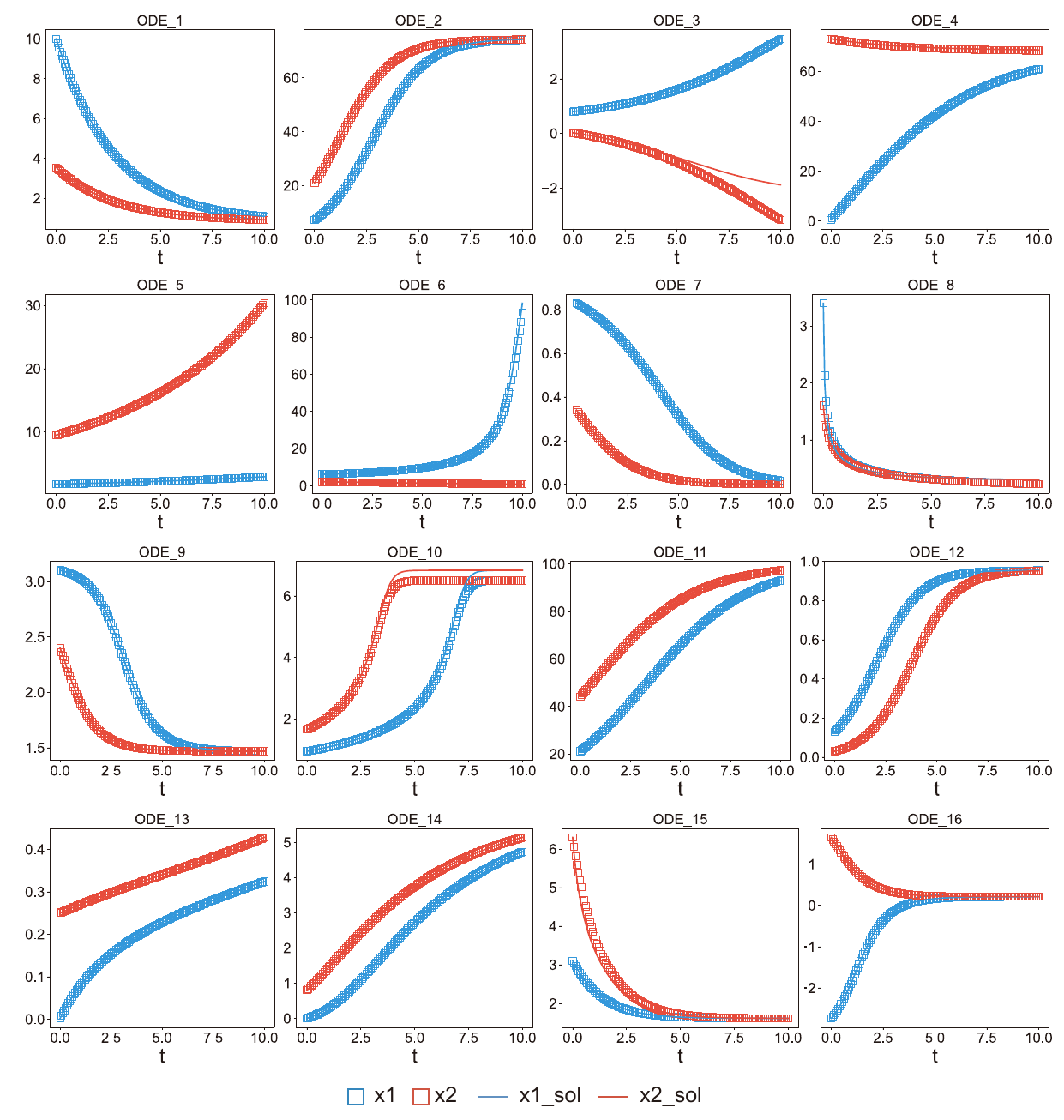}
\caption{Predicted solution trajectories on training and test sets.}
\label{fig:odesolution}
\end{figure}

\begin{table}[htb]
    \centering
    \caption{Discovered results and the reconstruction and generalization performance on the Strogatz dataset. We ran the experiments three times and presented the best expression for each ODE.}
 \resizebox{\textwidth}{!}{
    \begin{tabular}{ccccc}
    \toprule
        Benchmark& Discovered form & Parameters &$R^2$(train)  & $R^2$(test) \\
        \midrule
ODE-1 & $c_{0} + c_{1} x$ & [-0.3608,0.3031]& 0.999 & 0.999 \\\hline
ODE-2 & $c_{0} x^2 +c_{1} x$& [-0.0106,0.7899]&0.999 & 0.999\\\hline
ODE-3  & $c_{0} \sin{x} + c_{1} x^2 + c_{2} e^{c_{3}x}\sin{x} + c_{4}$ & [0.219,0.0563,0.0024,1.1,-0.1145] &0.999 & 0.727\\\hline
ODE-4 & $c_{0} x^2 + c_{1} $ &[-0.0021,9.8098] & 0.999 & 0.999 \\\hline
ODE-5  & $c_{0} x \log{\left(c_{1} x \right)}$ & [0.032,2.2901] &0.999& 0.973 \\\hline
ODE-6 & $c_{0} x^3 + c_{1} x^2+ c_{2} x$ & [-0.00024,0.033,-0.1408] & 0.996 & 0.999 \\\hline
ODE-7  & $c_{0}*x^2 -c_{1}*x*sin^2(x) -c_{2}*sin(x)*cos(x)$ & [1.2539,-1.2231,-0.7192] & 0.999 &0.999\\\hline
ODE-8  & $c_{0} x^3 + c_{1}$ & [-1.2554,0.0318] &0.979& 0.958 \\\hline
ODE-9 &  $c_{0} \sin{\left(x \right)} + c_{1} \sin{\left(x \right)} \cos{\left(x \right)}$ & [-0.0981, 0.9511] & 0.999 & 0.999 \\\hline
ODE-10 & $c_{0} x^5 + c_{1} x^3+ c_{2} x$ & [-0.0009,0.0399,0.1004] & 0.992 & 0.978\\\hline
ODE-11 &  $c_{0} x^2 + c_{1} x + c_{2}$ & [-0.004,0.3976,-0.0293]& 0.999 & 0.999\\\hline
ODE-12 & $c_{0}  \sin^2{\left(x \right)} + c_{1} x + c_{2} \cos{\left(x \right)} + c_{3}$& [0.464,0.907,2.7834,-2.7836]& 0.999 & 0.999\\\hline
ODE-13 & $c_{0}*exp(c_{1}*x)-c_{2}*sin(x)/x+c_{3}$ & [-0.2779,2.0, 9.7688,10.1468] & 0.999 & 0.999 \\\hline
ODE-14 & $c_{0} - c_{1} x - c_{2} e^{-x}$ & [1.1998,0.2,-0.9998] & 0.999 & 0.999 \\\hline
ODE-15 & $c_{0} x^2 + c_{1} x + c_{2}\sin{\left(x \right)} + c_{3}$ & [-0.1682,-0.2768,-0.5337,1.4144] & 0.999 & 0.977 \\\hline
ODE-16  & $c_{0} - c_{1}\sin{\left(x \right)}$ & [0.21,-0.9995] & 0.999 & 0.999 \\

    \bottomrule
    \end{tabular}
}
    
    \label{table:odebench}
\end{table}
In this section, we tested our framework on 16 one-dimensional ODEs in a comprehensive benchmark named ODE-bench~\cite{odeformer}, which has been utilized to describe real-world phenomena by Strogatz~\cite{strogatz2018nonlinear}.
The equation information is listed in Appendix~\ref{ode_info}. Each equation contains two sets of trajectories with two different initial conditions. We used one set of trajectory data as training data and searched to find the optimal $\hat \mathcal{F}$. During the evaluation process, we consider the solution trajectory of the ODE associated with $\mathcal{F}$ as predicted results and utilize the $R^2$ score as the evaluation criterion to measure the fitting accuracy in comparison to the actual trajectory. The $R^2$ value on the training set represents the accuracy of the reconstruction, while the $R^2$ value on the test set with a new initial condition signifies the generalization performance. As shown in Table~\ref{table:odebench}, we repeated the search process for each equation three times and reported the best results among them. It can be seen that in our framework, the percentage of equations with $R^2$ greater than 0.99 on the training set is 93.75\% (15/16), and on the test set, the percentage of equations with $R^2$  greater than 0.99 is 68.75\%. Equations with $R^2$ greater than 0.9 exceeded 90\% on both the training and test sets. Fig.~\ref{fig:odesolution} presents detailed prediction results for each equation. Most solution trajectories are consistent with the true trajectories.

\begin{table}[htb]

\caption{Evaluation on 16 ODEs with different methods and LLM backbones. We counted the number of equations discovered by different methods and LLMs that meet the corresponding $R^2$ criteria.  When the $R^2$ value is negative, or the solution trajectories of the ODE associated with $\hat \mathcal{ F}$ exhibit numerical overflow, we deem the discovered equation as "Invalid".}
\label{tabble:ode-compare}
 \resizebox{\textwidth}{!}{
\begin{tabular}{ccccccccc}
\hline

\multicolumn{2}{c}{\multirow{2}{*}{Methods}} & \multicolumn{3}{c}{Training set}               & \multicolumn{3}{c}{Testing set}                   & \multirow{2}{*}{Symbolically correct} \\
\multicolumn{2}{c}{}                         & $R^2$\textgreater{}0.99 & $R^2$\textgreater{}0.9 & Invaid & $R^2$\textgreater{}0.99 & $R^2$\textgreater{}0.9 & Invaid &                                       \\ \hline
\multicolumn{2}{c}{PySR~\cite{pysr}}                & \textbf{15 (93.75\%) }                 & 15 (93.75\%)            &1 (6.25\%) & 10 (62.5\%)   & 12 (75\%)                            & 4 (25\%)  & 6 (37.5\%)                                 \\ 
\multicolumn{2}{c}{ODEformer~\cite{odeformer}}                & 11 (68.75\%)                   & 14(87.5\%)               & 2 (12.5\%)   & 6 (37.5\%)                 & 9 (56.25\%)               & 5 (31.25\%)  & 3 (18.75\%)                                 \\ 
\hline
\multicolumn{1}{c|}{\multirow{3}{*}{Ours}}   & Llama2 7B   & 11 (68.75\%)                & 12 (75\%)                & 4 (25\%)    & 8 (50\%)                   & 11 (68.75\%)               & 4 (25\%)    &5 (31.25\%)                                \\
            \multicolumn{1}{c|}{} & GPT-3.5-turbo    & 13 (81.25\%)                & 13 (81.25\%)           &  3 (18.75\%) & \textbf{12 (75\%)}                   & 13 (81.25\%)              & 3 (18.75\%)  & \textbf{8 (50\%)}                                    \\
                                 \multicolumn{1}{c|}{} & GPT-4    &  \textbf{15 (93.75\%)}                 & \textbf{16 (100\%)}                    &  \textbf{0 (0\%)}   & 11 (68.75\%)     &    \textbf{15 (93.75\%)}             & \textbf{0 (0\%)}      &  \textbf{8 (50\%)}                                \\ \hline
\end{tabular}
}
\end{table}

\subsubsection{Comparisons with SR benchmarks and ablation studies}
We conducted a further comparison of the performance of SR benchmarks. Two symbolic regression benchmarks are utilized as the baseline models, including PySR~\cite{pysr} and ODEformer~\cite{odeformer}. PySR is a practical and high-performance library for symbolic regression, based on a multi-population evolutionary algorithm. PySR is designed for single-instance datasets and has been broadly adopted for interpretable symbolic discovery. ODEformer is based on pretrained transformers and achieves SOTA on ODEbench’s datasets. The aforementioned methods were tested according to their default hyperparameter configurations. PySR is implemented with a population size of 40 (the number of populations running) and total iterations of 40. ODEformer is implemented with a beam size of 20 and a temperature of 0.1.

As illustrated in Table~\ref{table:odebench}, our framework demonstrates equivalent reconstruction performance to the aforementioned methods designed for symbolic regression, while exhibiting superior generalization capabilities and usability. Note that PySR, which is based on evolutionary search, possesses powerful search capabilities and can discover equations that are numerically accurate on the training dataset. However, these discovered equations tend to have more complex structures and, consequently, are susceptible to exhibiting poor generalization performance. 

Additionally, we performed a comparative analysis of the performance of LLM backbones with varying parameter sizes and language capabilities on this task. We used the open-source large language model Llama2 with 7B parameters~\cite{touvron2023llama} and more advanced models, including GPT-3.5-turbo and GPT-4. The results indicate that as the capability of LLMs improves, the identified equations become relatively more accurate on both the training set and the test set. This can be primarily attributed to the fact that the capability of the LLM directly exerts a substantial influence on the generation and optimization of equations. On the one hand, large models with inferior capabilities may struggle to effectively understand and execute the provided instructions, such as the constraints we defined, leading to the generation of numerous invalid equations. Furthermore, they tend to fail to properly execute GA instructions and accurately perform crossover and mutation operations. Conversely, the model's reasoning capability directly influences its self-improvement optimization capabilities. It is also worth noting that as the capabilities and number of parameters of large models further increase, the gains in accuracy diminish, especially on test sets.

In addition, we further emphasize the efficiency of the proposed framework. The total running time is the product of the time per iteration and the total number of iterations. The time cost per iteration ranges from 10s to 40s under the default configuration and mainly includes the time spent on remotely accessing the LLM API and evaluating the feedback from the LLM, i.e., the generated equations. The time consumed for accessing the API interface exhibits a positive correlation with the number of samples generated and is roughly an order of magnitude larger than the time for evaluations. In practice, we can allocate separate processes to query the LLM in parallel. On one hand, this approach can reduce the total response time. On the other hand, we can increase the number of expressions generated in each iteration, which in turn contributes to more accurate samples for optimization and helps reduce the number of optimization iterations.

\section{Conclusion}
We introduce a novel equation discovery framework guided by LLMs. It aims to facilitate equation discovery across diverse domains, transcending the confines of specialist communities, and making LLM-guided discovery accessible to a broader range of users. The framework leverages the generation and reasoning capabilities of LLMs to automatically generate and optimize equations. We employ natural language-based prompts to guide LLMs in conducting iterative optimization using self-improvement and genetic algorithms. The results indicate that the two strategies exhibit a strong synergistic effect, effectively balancing exploration and exploitation. Experiments demonstrate that the proposed framework can discover effective and accurate equations (including both PDEs and ODEs), directly from data. More importantly, our framework achieved performance comparable to the SR benchmarks, and the discovered equations tend to be more physically reasonable and possess better generalization ability. The impact of LLM capabilities on the performance of mined equations is also thoroughly discussed. The proposed framework demonstrates great potential in applying LLMs to scientific discovery and providing interpretability for complex physical phenomena.

The current framework has the potential for enhancement in two aspects. First, the training corpus of LLMs encompasses descriptions of diverse intricate physical processes and phenomena; therefore, providing a depiction of the nonlinear system to be mined and information on physical variables could be crucial for LLMs to generate rational equation forms. Further exploration is required to design natural language-based prompts that incorporate prior knowledge to effectively reduce the search space and enhance search efficiency and equation mining precision. Second, additional experimental validation is necessary to determine the robustness of equation mining using LLMs, particularly in scenarios with sparse and noisy observations. Integrating more efficient evaluation techniques might be essential for addressing more intricate situations.

\section*{Acknowledgments}
This work was supported and partially funded by the National Center for Applied Mathematics Shenzhen (NCAMS), the Shenzhen Key Laboratory of Natural Gas Hydrates (Grant No. ZDSYS20200421111201738), the SUSTech – Qingdao New Energy Technology Research Institute, the China Meteorological
Administration Climate Change Special Program (CMA-CCSP) (Grant No. QBZ202316),
 and the National Natural Science Foundation of China (Grant No. 62106116).

\section*{Code and data availability}
The implementation details of the whole process and relevant data are available on GitHub at https://github.com/menggedu/EDL.

\appendix
\section{Prompts}
\label{append:prompt}

\begin{figure}[!ht]
\centering
\includegraphics[width=0.8\textwidth]{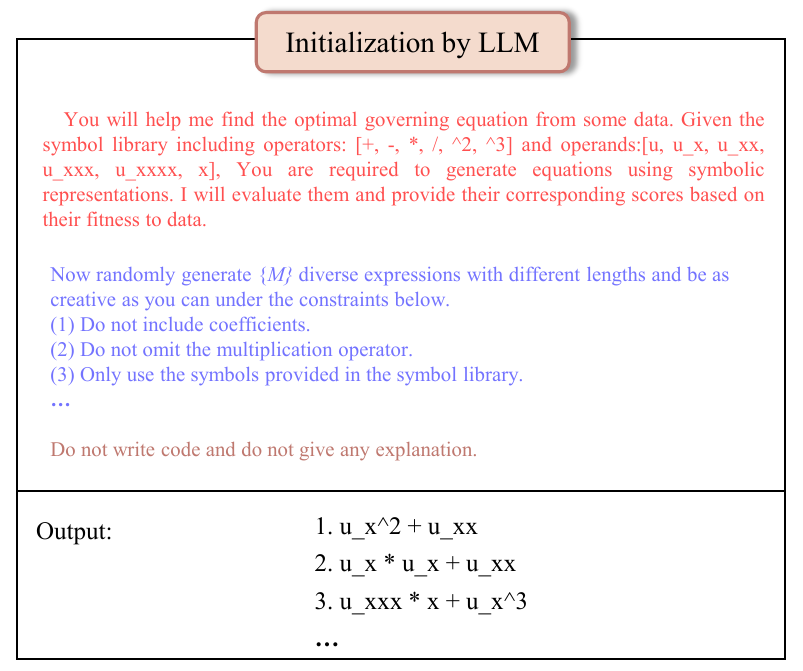}
\caption{Prompt utilized in initialization.}
\label{fig:init_prompt}
\end{figure}

\begin{figure}[!ht]
\centering
\includegraphics[width=0.8\textwidth]{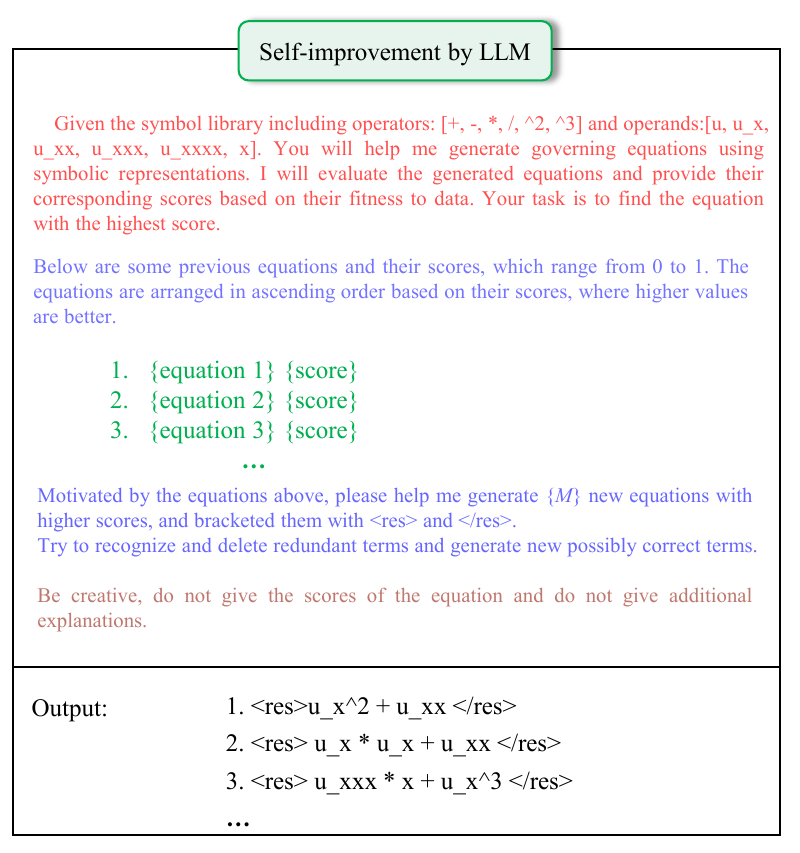}
\caption{Prompt utilized for self-improvement.}
\label{fig:SI_prompt}
\end{figure}

\begin{figure}[!ht]
\centering
\includegraphics[width=0.8\textwidth]{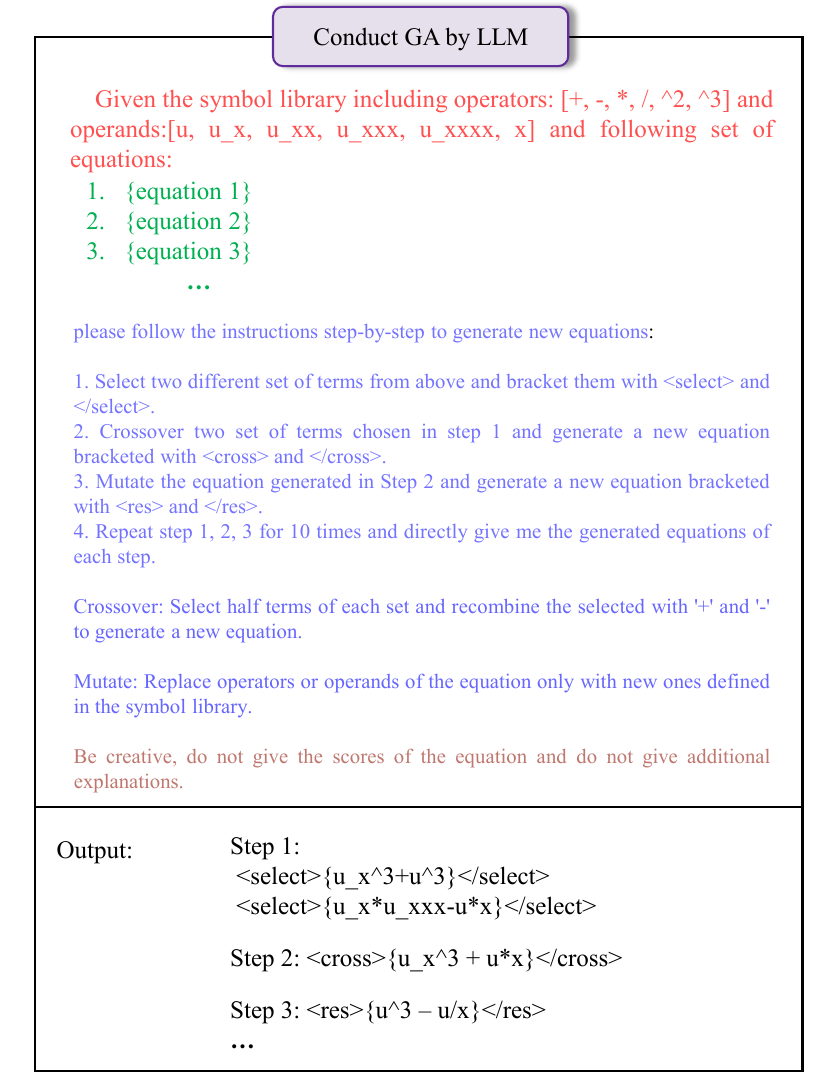}
\caption{Prompt utilized for evolutionary search.}
\label{fig:GA_prompt}
\end{figure}

The prompts utilized for initialization, self-improvement, and GA are shown in Fig.~\ref{fig:init_prompt}, Fig.~\ref{fig:SI_prompt}, and Fig.~\ref{fig:GA_prompt}, respectively. Different colors represent different components in the prompt. Specifically, red represents task descriptions, blue represents instructions, green represents historical samples, and brown represents other hints and constraints.

\section{ODE datasets}
The ODE datasets are selected from ODEBench~\cite{odeformer} with different skeletons. The information for each ode is shown in Table~\ref{table:ode-bench_sup}.
\label{ode_info}

\begin{table}[htb]
    \centering
    \caption{Scalar ODEs from ODEBench~\cite{odeformer}.}
    \label{table:ode-bench_sup}
\scalebox{0.8}{
    \begin{tabular}{l|p{6cm}|l|c|c}
    \toprule
        ID & System description & Equation & Parameters & initial conditions\\
        \midrule
1 & RC-circuit (charging capacitor) & $\frac{c_{0} - \frac{x_{0}}{c_{1}}}{c_{2}}$ & 0.7, 1.2, 2.31 & [10.0], [3.54]\\\hline
2 & Population growth with carrying capacity & $c_{0} x_{0} \cdot \left(1 - \frac{x_{0}}{c_{1}}\right)$ & 0.79, 74.3 & [7.3], [21.0]\\\hline
3 & RC-circuit with non-linear resistor (charging capacitor) & $-0.5 + \frac{1}{e^{c_{0} - \frac{x_{0}}{c_{1}}} + 1}$ & 0.5, 0.96 & [0.8], [0.02]\\\hline
4 & Velocity of a falling object with air resistance & $c_{0} - c_{1} x_{0}^{2}$ & 9.81, 0.0021175 & [0.5], [73.0]\\\hline
5 & Gompertz law for tumor growth & $c_{0} x_{0} \log{\left(c_{1} x_{0} \right)}$ & 0.032, 2.29 & [1.73], [9.5]\\\hline
6 & Logistic equation with Allee effect & $c_{0} x_{0} \left(-1 + \frac{x_{0}}{c_{2}}\right) \left(1 - \frac{x_{0}}{c_{1}}\right)$ & 0.14, 130.0, 4.4 & [6.123], [2.1]\\\hline
7 & Refined language death model for two languages & $c_{0} x_{0}^{c_{1}} \cdot \left(1 - x_{0}\right) - x_{0} \cdot \left(1 - c_{0}\right) \left(1 - x_{0}\right)^{c_{1}}$ & 0.2, 1.2 & [0.83], [0.34]\\\hline
8 & Overdamped bead on a rotating hoop & $c_{0} \left(c_{1} \cos{\left(x_{0} \right)} - 1\right) \sin{\left(x_{0} \right)}$ & 0.0981, 9.7 & [3.1], [2.4]\\\hline
9 & Budworm outbreak with predation (dimensionless) & $c_{0} x_{0} \cdot \left(1 - \frac{x_{0}}{c_{1}}\right) - \frac{x_{0}^{2}}{x_{0}^{2} + 1}$ & 0.4, 95.0 & [44.3], [4.5]\\\hline
10 & Landau equation (typical time scale tau = 1) & $c_{0} x_{0} - c_{1} x_{0}^{3} - c_{2} x_{0}^{5}$ & 0.1, -0.04, 0.001 & [0.94], [1.65]\\\hline
11 & Improved logistic equation with harvesting/fishing & $c_{0} x_{0} \cdot \left(1 - \frac{x_{0}}{c_{1}}\right) - \frac{c_{2} x_{0}}{c_{3} + x_{0}}$ & 0.4, 100.0, 0.24, 50.0 & [21.1], [44.1]\\\hline
12 & Improved logistic equation with harvesting/fishing (dimensionless) & $- \frac{c_{0} x_{0}}{c_{1} + x_{0}} + x_{0} \cdot \left(1 - x_{0}\right)$ & 0.08, 0.8 & [0.13], [0.03]\\\hline
13 & Autocatalytic gene switching (dimensionless) & $c_{0} - c_{1} x_{0} + \frac{x_{0}^{2}}{x_{0}^{2} + 1}$ & 0.1, 0.55 & [0.002], [0.25]\\\hline
14 & Dimensionally reduced SIR infection model for dead people (dimensionless) & $c_{0} - c_{1} x_{0} - e^{- x_{0}}$ & 1.2, 0.2 & [0.0], [0.8]\\\hline
15 & Hysteretic activation of a protein expression (positive feedback, basal promoter expression) & $c_{0} + \frac{c_{1} x_{0}^{5}}{c_{2} + x_{0}^{5}} - c_{3} x_{0}$ & 1.4, 0.4, 123.0, 0.89 & [3.1], [6.3]\\\hline
16 & Overdamped pendulum with constant driving torque/fireflies/Josephson junction (dimensionless) & $c_{0} - \sin{\left(x_{0} \right)}$ & 0.21 & [-2.74], [1.65]\\

    \bottomrule
    \end{tabular}
    }

\end{table}

\bibliographystyle{unsrt}  
\bibliography{references}

\begin{thebibliography}{10}

\bibitem{yeo2019deep}
Kyongmin Yeo and Igor Melnyk.
\newblock Deep learning algorithm for data-driven simulation of noisy dynamical system.
\newblock {\em Journal of Computational Physics}, 376:1212--1231, 2019.

\bibitem{kou2021data}
Jiaqing Kou and Weiwei Zhang.
\newblock Data-driven modeling for unsteady aerodynamics and aeroelasticity.
\newblock {\em Progress in Aerospace Sciences}, 125:100725, 2021.

\bibitem{zheng2020purely}
Gang Zheng, Xiaofeng Li, Rong-Hua Zhang, and Bin Liu.
\newblock Purely satellite data--driven deep learning forecast of complicated tropical instability waves.
\newblock {\em Science advances}, 6(29):eaba1482, 2020.

\bibitem{ISgep}
Samaneh~Sadat {Mousavi Astarabadi} and Mohammad~Mehdi Ebadzadeh.
\newblock Genetic programming performance prediction and its application for symbolic regression problems.
\newblock {\em Information Sciences}, 502:418--433, 2019.

\bibitem{2022pinnreview2}
George~Em Karniadakis, Ioannis~G Kevrekidis, Lu~Lu, Paris Perdikaris, Sifan Wang, and Liu Yang.
\newblock Physics-informed machine learning.
\newblock {\em Nature Reviews Physics}, 3(6):422--440, 2021.

\bibitem{lu2021physics}
Lu~Lu, Raphael Pestourie, Wenjie Yao, Zhicheng Wang, Francesc Verdugo, and Steven~G Johnson.
\newblock Physics-informed neural networks with hard constraints for inverse design.
\newblock {\em SIAM Journal on Scientific Computing}, 43(6):B1105--B1132, 2021.

\bibitem{chen_inte}
Yuntian Chen and Dongxiao Zhang.
\newblock Integration of knowledge and data in machine learning.
\newblock {\em arXiv preprint arXiv:2202.10337}, 2022.

\bibitem{schmidt}
Michael Schmidt and Hod Lipson.
\newblock Distilling free-form natural laws from experimental data.
\newblock {\em Science}, 324(5923):81--85, 2009.

\bibitem{2019pnasDD}
Kathleen Champion, Bethany Lusch, J~Nathan Kutz, and Steven~L Brunton.
\newblock Data-driven discovery of coordinates and governing equations.
\newblock {\em Proceedings of the National Academy of Sciences}, 116(45):22445--22451, 2019.

\bibitem{sindy}
Steven~L Brunton, Joshua~L Proctor, and J~Nathan Kutz.
\newblock Discovering governing equations from data by sparse identification of nonlinear dynamical systems.
\newblock {\em Proc. Natl. Acad. Sci.}, 113(15):3932--3937, 2016.

\bibitem{sindy_bvp}
Daniel~E Shea, Steven~L Brunton, and J~Nathan Kutz.
\newblock Sindy-bvp: Sparse identification of nonlinear dynamics for boundary value problems.
\newblock {\em Phys. Rev. Res.}, 3(2):023255, 2021.

\bibitem{weak_sindy}
Daniel~A Messenger and David~M Bortz.
\newblock Weak sindy for partial differential equations.
\newblock {\em J. Comput. Phys.}, 443:110525, 2021.

\bibitem{sindy_pi}
Kadierdan Kaheman, J~Nathan Kutz, and Steven~L Brunton.
\newblock Sindy-pi: a robust algorithm for parallel implicit sparse identification of nonlinear dynamics.
\newblock {\em Proceedings of the Royal Society A}, 476(2242):20200279, 2020.

\bibitem{ensemble_sindy}
Urban Fasel, J~Nathan Kutz, Bingni~W Brunton, and Steven~L Brunton.
\newblock Ensemble-sindy: Robust sparse model discovery in the low-data, high-noise limit, with active learning and control.
\newblock {\em Proceedings of the Royal Society A}, 478(2260):20210904, 2022.

\bibitem{eql}
Georg Martius and Christoph~H Lampert.
\newblock Extrapolation and learning equations.
\newblock {\em arXiv preprint arXiv:1610.02995}, 2016.

\bibitem{eql_divide}
Subham Sahoo, Christoph Lampert, and Georg Martius.
\newblock Learning equations for extrapolation and control.
\newblock In {\em International Conference on Machine Learning}, pages 4442--4450. PMLR, 2018.

\bibitem{kamienny2023generativesr}
Pierre-Alexandre Kamienny, Guillaume Lample, Sylvain Lamprier, and Marco Virgolin.
\newblock Deep generative symbolic regression with monte-carlo-tree-search.
\newblock In {\em International Conference on Machine Learning}, pages 15655--15668. PMLR, 2023.

\bibitem{valipour2021symbolicgpt}
Mojtaba Valipour, Bowen You, Maysum Panju, and Ali Ghodsi.
\newblock Symbolicgpt: A generative transformer model for symbolic regression.
\newblock {\em arXiv preprint arXiv:2106.14131}, 2021.

\bibitem{li2022transformer}
Wenqiang Li, Weijun Li, Linjun Sun, Min Wu, Lina Yu, Jingyi Liu, Yanjie Li, and Songsong Tian.
\newblock Transformer-based model for symbolic regression via joint supervised learning.
\newblock In {\em The Eleventh International Conference on Learning Representations}, 2022.

\bibitem{dsr}
Brenden~K Petersen, Mikel~Landajuela Larma, Terrell~N Mundhenk, Claudio~Prata Santiago, Soo~Kyung Kim, and Joanne~Taery Kim.
\newblock Deep symbolic regression: Recovering mathematical expressions from data via risk-seeking policy gradients.
\newblock In {\em International Conference on Learning Representations}, 2021.

\bibitem{sun2022symbolic}
Fangzheng Sun, Yang Liu, Jian-Xun Wang, and Hao Sun.
\newblock Symbolic physics learner: Discovering governing equations via monte carlo tree search.
\newblock In {\em International Conference on Learning Representations}, 2023.

\bibitem{PRRDISCOVER}
Mengge Du, Yuntian Chen, and Dongxiao Zhang.
\newblock Discover: Deep identification of symbolically concise open-form partial differential equations via enhanced reinforcement learning.
\newblock {\em Phys. Rev. Res.}, 6:013182, Feb 2024.

\bibitem{sga_pde}
Yuntian Chen, Yingtao Luo, Qiang Liu, Hao Xu, and Dongxiao Zhang.
\newblock Symbolic genetic algorithm for discovering open-form partial differential equations (sga-pde).
\newblock {\em Phys. Rev. Res.}, 4(2):023174, 2022.

\bibitem{Rdiscover}
Mengge Du, Yuntian Chen, Longfeng Nie, Siyu Lou, and Dongxiao Zhang.
\newblock Physics-constrained robust learning of open-form partial differential equations from limited and noisy data.
\newblock {\em Physics of Fluids}, 36(5), 2024.

\bibitem{ISgenetic}
Samaneh Sadat~Mousavi Astarabadi and Mohammad~Mehdi Ebadzadeh.
\newblock Genetic programming performance prediction and its application for symbolic regression problems.
\newblock {\em Information Sciences}, 502:418--433, 2019.

\bibitem{gp_sr2}
Samaneh Sadat~Mousavi Astarabadi and Mohammad~Mehdi Ebadzadeh.
\newblock Genetic programming performance prediction and its application for symbolic regression problems.
\newblock {\em Inf. Sci.}, 502:418--433, 2019.

\bibitem{gp_sr3}
Sheng Sun, Runhai Ouyang, Bochao Zhang, and Tong-Yi Zhang.
\newblock Data-driven discovery of formulas by symbolic regression.
\newblock {\em MRS Bull.}, 44(7):559--564, 2019.

\bibitem{LLM_utilize1}
Bonan Min, Hayley Ross, Elior Sulem, Amir Pouran~Ben Veyseh, Thien~Huu Nguyen, Oscar Sainz, Eneko Agirre, Ilana Heintz, and Dan Roth.
\newblock Recent advances in natural language processing via large pre-trained language models: A survey.
\newblock {\em ACM Computing Surveys}, 56(2):1--40, 2023.

\bibitem{LLM_utilize2}
Claudia~E Haupt and Mason Marks.
\newblock Ai-generated medical advice—gpt and beyond.
\newblock {\em Jama}, 329(16):1349--1350, 2023.

\bibitem{LLM_utilize3}
Augustin Lecler, Lo{\"\i}c Duron, and Philippe Soyer.
\newblock Revolutionizing radiology with gpt-based models: current applications, future possibilities and limitations of chatgpt.
\newblock {\em Diagnostic and Interventional Imaging}, 104(6):269--274, 2023.

\bibitem{LLM_mathematicaldiscover}
Bernardino Romera-Paredes, Mohammadamin Barekatain, Alexander Novikov, Matej Balog, M~Pawan Kumar, Emilien Dupont, Francisco~JR Ruiz, Jordan~S Ellenberg, Pengming Wang, Omar Fawzi, et~al.
\newblock Mathematical discoveries from program search with large language models.
\newblock {\em Nature}, 625(7995):468--475, 2024.

\bibitem{AEL_evo}
Fei Liu, Xialiang Tong, Mingxuan Yuan, Xi~Lin, Fu~Luo, Zhenkun Wang, Zhichao Lu, and Qingfu Zhang.
\newblock An example of evolutionary computation + large language model beating human: Design of efficient guided local search, 2024.

\bibitem{wang2023codet5+}
Yue Wang, Hung Le, Akhilesh~Deepak Gotmare, Nghi~DQ Bui, Junnan Li, and Steven~CH Hoi.
\newblock Codet5+: Open code large language models for code understanding and generation.
\newblock {\em arXiv preprint arXiv:2305.07922}, 2023.

\bibitem{liullm_es_optimizer}
Shengcai Liu, Caishun Chen, Xinghua Qu, Ke~Tang, and Yew-Soon Ong.
\newblock Large language models as evolutionary optimizers.
\newblock {\em arXiv preprint arXiv:2310.19046}, 2023.

\bibitem{ISbrence2023}
Jure Brence, Sa{\v{s}}o D{\v{z}}eroski, and Ljup{\v{c}}o Todorovski.
\newblock Dimensionally-consistent equation discovery through probabilistic attribute grammars.
\newblock {\em Information Sciences}, 632:742--756, 2023.

\bibitem{cfg_parkes2008concise}
Alan~P Parkes.
\newblock {\em A concise introduction to languages and machines}.
\newblock Springer Science \& Business Media, 2008.

\bibitem{dsr_gp}
Terrell Mundhenk, Mikel Landajuela, Ruben Glatt, Claudio~P Santiago, Daniel faissol, and Brenden~K Petersen.
\newblock Symbolic regression via deep reinforcement learning enhanced genetic programming seeding.
\newblock In {\em Advances in Neural Information Processing Systems}, volume~34, pages 24912--24923, 2021.

\bibitem{gp_sr4}
Maryam~Amir Haeri, Mohammad~Mehdi Ebadzadeh, and Gianluigi Folino.
\newblock Statistical genetic programming for symbolic regression.
\newblock {\em Appl. Soft Comput.}, 60:447--469, 2017.

\bibitem{biggio2021neural}
Luca Biggio, Tommaso Bendinelli, Alexander Neitz, Aurelien Lucchi, and Giambattista Parascandolo.
\newblock Neural symbolic regression that scales.
\newblock In {\em International Conference on Machine Learning}, pages 936--945. Pmlr, 2021.

\bibitem{vastl2024symformer}
Martin Vastl, Jon{\'a}{\v{s}} Kulh{\'a}nek, Ji{\v{r}}{\'\i} Kubal{\'\i}k, Erik Derner, and Robert Babu{\v{s}}ka.
\newblock Symformer: End-to-end symbolic regression using transformer-based architecture.
\newblock {\em IEEE Access}, 2024.

\bibitem{odeformer}
Stéphane d'Ascoli, Sören Becker, Alexander Mathis, Philippe Schwaller, and Niki Kilbertus.
\newblock Odeformer: Symbolic regression of dynamical systems with transformers, 2023.

\bibitem{LLM_utilize4}
Daniel~Martin Katz, Michael~James Bommarito, Shang Gao, and Pablo Arredondo.
\newblock Gpt-4 passes the bar exam.
\newblock {\em Philosophical Transactions of the Royal Society A}, 382(2270):20230254, 2024.

\bibitem{Yang_selfoptimize}
Chengrun Yang, Xuezhi Wang, Yifeng Lu, Hanxiao Liu, Quoc~V Le, Denny Zhou, and Xinyun Chen.
\newblock Large language models as optimizers.
\newblock In {\em The Twelfth International Conference on Learning Representations}, 2024.

\bibitem{guo_selfoptimize}
Pei-Fu Guo, Ying-Hsuan Chen, Yun-Da Tsai, and Shou-De Lin.
\newblock Towards optimizing with large language models, 2023.

\bibitem{liufeiMOEO_LLM}
Fei Liu, Xi~Lin, Zhenkun Wang, Shunyu Yao, Xialiang Tong, Mingxuan Yuan, and Qingfu Zhang.
\newblock Large language model for multi-objective evolutionary optimization, 2024.

\bibitem{promot_evo}
Qingyan Guo, Rui Wang, Junliang Guo, Bei Li, Kaitao Song, Xu~Tan, Guoqing Liu, Jiang Bian, and Yujiu Yang.
\newblock Connecting large language models with evolutionary algorithms yields powerful prompt optimizers.
\newblock In {\em The Twelfth International Conference on Learning Representations}, 2024.

\bibitem{Lan_LLMea_game}
Pier~Luca Lanzi and Daniele Loiacono.
\newblock Chatgpt and other large language models as evolutionary engines for online interactive collaborative game design.
\newblock In {\em Proceedings of the Genetic and Evolutionary Computation Conference}, GECCO '23, page 1383–1390, New York, NY, USA, 2023. Association for Computing Machinery.

\bibitem{LLM4ES}
Robert~Tjarko Lange, Yingtao Tian, and Yujin Tang.
\newblock Large language models as evolution strategies, 2024.

\bibitem{pde_find}
Samuel~H Rudy, Steven~L Brunton, Joshua~L Proctor, and J~Nathan Kutz.
\newblock Data-driven discovery of partial differential equations.
\newblock {\em Sci. Adv.}, 3(4):e1602614, 2017.

\bibitem{meurer2017sympy}
Aaron Meurer, Christopher~P Smith, Mateusz Paprocki, Ond{\v{r}}ej {\v{C}}ert{\'\i}k, Sergey~B Kirpichev, Matthew Rocklin, AMiT Kumar, Sergiu Ivanov, Jason~K Moore, Sartaj Singh, et~al.
\newblock Sympy: symbolic computing in python.
\newblock {\em PeerJ Computer Science}, 3:e103, 2017.

\bibitem{BRENCE_PROGED}
Jure Brence, Ljupčo Todorovski, and Sašo Džeroski.
\newblock Probabilistic grammars for equation discovery.
\newblock {\em Knowledge-Based Systems}, 224:107077, 2021.

\bibitem{bfgs}
John~D Head and Michael~C Zerner.
\newblock A broyden—fletcher—goldfarb—shanno optimization procedure for molecular geometries.
\newblock {\em Chemical physics letters}, 122(3):264--270, 1985.

\bibitem{ga1}
Tom~V Mathew.
\newblock Genetic algorithm.
\newblock {\em Report submitted at IIT Bombay}, page~53, 2012.

\bibitem{ga2}
Stephanie Forrest.
\newblock Genetic algorithms.
\newblock {\em ACM Comput. Surv.}, 28(1):77--80, 1996.

\bibitem{strogatz2018nonlinear}
Steven~H Strogatz.
\newblock {\em Nonlinear dynamics and chaos: with applications to physics, biology, chemistry, and engineering}.
\newblock CRC press, 2018.

\bibitem{pysr}
Miles Cranmer.
\newblock Interpretable machine learning for science with pysr and symbolicregression.jl, 2023.

\bibitem{touvron2023llama}
Hugo Touvron, Louis Martin, Kevin Stone, Peter Albert, Amjad Almahairi, Yasmine Babaei, Nikolay Bashlykov, Soumya Batra, Prajjwal Bhargava, Shruti Bhosale, et~al.
\newblock Llama 2: Open foundation and fine-tuned chat models.
\newblock {\em arXiv preprint arXiv:2307.09288}, 2023.

\end{thebibliography}

\end{document}